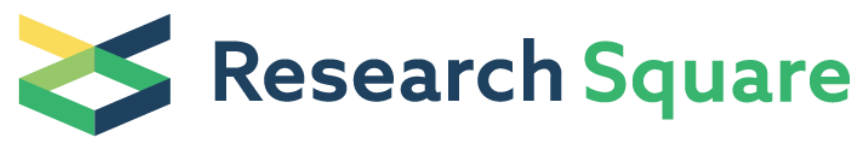

# Developing and Validating the Spanish Version of the Large Language Models Dependency Scale (LLM-D12-SP)

**Tran Gia Bao**
VinUniversity
**Mo El-Haj**
Lancaster University
**Sameha Al-Shakhsi**
Hamad Bin Khalifa University
**Antonio Garcia-Cabot**
Universidad de Alcalá
**Raian Ali**
Hamad Bin Khalifa University
**Ala Yankouskaya**
ayankouskaya@bournemouth.ac.uk
Bournemouth University

**Additional Declarations:** No competing interests reported.

## Abstract

There is a growing need for reliable and culturally validated instruments to assess psychological dependency on large language models (LLMs), particularly as LLMs are increasingly used for task execution, decision-making, and communication in organizational and work-related settings. This need is especially relevant for Spanish-speaking populations, where LLM adoption is rapidly expanding, yet validated psychometric tools remain scarce. The present study reports the first validation of the Spanish version of the Large Language Model Dependency Scale (LLM-D12-SP), extending prior validations conducted in English- and Arabic-speaking samples. The LLM-D12 is a two-dimensional instrument assessing Instrumental Dependency (reliance on LLMs for performing tasks and supporting decisions) and Relationship Dependency (psychological reliance on LLMs for companionship and social interaction). A total of 386 Spanish-speaking participants (M = 28.0 years, SD = 6.1; 55% male) completed the LLM-D12-SP. Confirmatory factor analysis supported the original two-factor structure. The scale demonstrated good internal consistency (Cronbach's α = 0.89 total; 0.86 Instrumental; 0.85 Relationship). Discriminant validity analyses indicated that the two subscales represent related but distinct constructs. External validation showed that both dependency dimensions were positively associated with internet addiction and perceived trustworthiness of LLMs, while showing weak or no association with need for cognition. Together with prior English and Arabic validations, these findings establish cross-linguistic support for the scale's structure and provide a psychometrically sound tool for investigating psychological aspects of LLM use in organizational contexts.

## 1. Introduction

The rapid advancement of Large Language Models (LLMs) over the past five years has fundamentally reshaped human interaction with digital technologies. These systems now serve as cognitive partners across education [1], healthcare [2] law [3] finance [4], and public policy [5], offering unprecedented support in information synthesis, decision-making, creative ideation, and problem-solving. Their integration into daily life extends beyond professional domains: individuals routinely consult LLMs to draft communications, evaluate options, manage uncertainty, and even seek emotional reassurance. This widespread adoption reflects LLMs' capacity to augment human capabilities, enhance efficiency, and make easy access to knowledge.

This integration carries a dual perspective. As LLMs transition from occasional tools to habitual cognitive reference points, concerns have emerged regarding their impact on human autonomy [6] critical thinking [7] and self-efficacy [8]. When users consistently delegate reasoning, judgment, or emotional processing to LLMs, they may experience a gradual erosion of confidence in their own capacities, particularly when operating without LLM support [9]. This shift marks more than a change in tool use; it signals an emerging form of behavioral reliance that encompasses both instrumental and relational dimensions. Instrumental reliance refers to the use of LLMs for cognitive offloading and task execution, whereas instrumental dependence arises when users come to rely on LLMs as necessary supports for performing tasks they could previously complete independently [10][11]. Relational reliance involves engaging with LLMs for social or affective interaction, while relational dependence reflects a tendency to prioritize LLM-mediated interaction over human communication by attributing empathy, companionship, or quasi-human qualities to the system, sharing private details and feeling less need to talk to people [12].

The dual-layered dependence creates conditions in which LLMs are not merely used but psychologically integrated into users' self-concept and problem-solving identity [13]. Over time, the habitual offloading of judgment to an AI that responds with human-like fluency and apparent understanding can recalibrate users' internal benchmarks for competence, leading them to perceive their unaided cognition as insufficient. In this way, LLM dependency emerges not from passive overuse, but from an active, often unconscious, reconfiguration of the self-in-relation-to-technology, a shift where LLMs becomes a constitutive element of one's cognitive [14] and emotional scaffolding [15][16].

This evolving dynamic aligns with broader theoretical frameworks in human–technology interaction. Technologies can reshape users' cognitive processes, as described in the concept of cognitive artifacts, while uncritical trust in algorithmic outputs has been cautioned against in work on automation bias [17][18]. Recent empirical findings corroborate these concerns: users often accept LLM-generated content as authoritative despite factual inaccuracies, show reduced persistence in unaided problem-solving after LLM collaboration [19], and report discomfort or diminished self-efficacy when AI tools are unavailable [20]. Critically, this pattern is not merely about frequency of use but about the internalization of LLM as a necessary component of cognitive and emotional functioning, a phenomenon increasingly described as LLM dependency [10][21][22].

## 1.2. Measuring LLM Dependency: current instruments and conceptual limitations

Several instruments have been developed to assess problematic or dependency-like patterns of generative AI use.

- Problematic ChatGPT Use Scale (PCUS; [23]. An 11-item measure grounded in DSM-5 addiction criteria, designed to capture cognitive, behavioral and emotional aspects of problematic ChatGPT use, including preoccupation, withdrawal, tolerance, loss of control, conflict and mood modification.
- Problematic Use of Conversational Artificial Intelligence Scale (PUCAI-6; [24]: A brief six-item scale adapted from the Bergen Social Media Addiction Scale [25] to index addictive use of conversational AI, with items reflecting core components such as salience, withdrawal and conflict. It has been widely used to examine links between problematic CAI use and social anxiety, loneliness and rumination.
- AI Chatbot Dependence Scale (AICD;[26]: An eight-item unidimensional instrument developed to assess dependence on AI chatbots, drawing on smartphone and social media addiction frameworks. It focuses on information-seeking, task execution and the overall intensity of chatbot use that begins to interfere with everyday functioning.
- Generative AI Dependency Scale [27]. An 11-item multidimensional scale that measures individual differences in dependency on generative AI tools across platforms. The final structure captures domains such as cognitive preoccupation with generative AI, negative consequences in social, academic or occupational functioning, and withdrawal-like discomfort when access is limited, and has been validated against motivational, behavioral and psychological correlates in community samples.

- LLMs Dependence Scale (LDS) [28]. An 18 -items scale assessing two dimensions – Existential Dependence and Functional Dependence. The item set captures domains reflecting cognitive and emotional reliance on LLMs, loss of autonomous problem-solving, and functional or psychological discomfort when LLM support is unavailable. The LDS development is explicitly grounded in addiction theory, integrated with human-computer trust frameworks, positioning LLM dependence as a progression from instrumental assistance to maladaptive reliance rather than mere high-frequency use.

These instruments represent an important first generation of tools for studying problematic engagement with generative AI. They show that LLM dependency patterns can be operationalized, demonstrate acceptable reliability, and link elevated scores to outcomes such as reduced wellbeing, lower critical thinking, and disruption to social or academic functioning. The emergence of these measures suggests that over-reliance on LLMs and chatbots is not always benign and can, for some users, be associated with more compulsive patterns of use.

At the same time, these scales are grounded largely in frameworks developed for substance and behavioral addictions, which foreground symptoms such as preoccupation, tolerance, loss of control and withdrawal. This orientation is well suited to detecting harmful overuse, but it captures only a subset of the ways in which people now integrate LLMs into everyday thinking, learning and interpersonal life. In practice, this direction in LLM dependency tends to emphasize LLMs as possible sources of compulsion, with less emphasis on their roles as cognitive tools and quasi-relational partners. In addition, current measures usually combine distinct forms of reliance into a single severity continuum. They do not separate reliance on LLMs for instrumental functions (e.g. planning, decision support, memory offloading) from reliance that reflects changes in self-confidence or agency (e.g. feeling unable to write or decide without the model), or from relational use (e.g. seeking comfort, validation or companionship). As a result, individuals who use LLMs intensively to increase efficiency and those who use them to cope with loneliness may obtain similar “problematic use” scores, even though the underlying purposes and contexts of use may diverge in important ways.

Recent advances in the measurement of LLM dependency have led to the development of the Large Language Model Dependency Scale (LLM-D12) [11], a theoretically grounded instrument designed to capture qualitatively distinct forms of reliance on AI systems [10]. The scale comprises two six-item subscales. Instrumental dependency captures the extent to which individuals integrate LLMs into cognitive tasks, judgement formation, and decision-making, whereas relational dependency reflects the degree to which LLMs are experienced as socially responsive and emotionally engaging. This two-factor structure demonstrated excellent internal consistency and good discriminant validity, with external validation supporting both the conceptual grounding of the measure and the distinction between its subscales. Importantly, the psychometric properties of the LLM-D12 are interpreted in line with the emerging view that dependency on LLMs does not inherently imply dysfunction, while still indexing forms of reliance that may become problematic in specific contexts. Validation in English [11] and Arabic [29] populations confirmed a robust two-factor structure, high internal consistency (Cronbach’s $\alpha > 0.85$), and strong discriminant validity.

To date, no validated version of the LLM-D12 exists for Spanish-speaking users, a significant gap given that Spanish is the native language of over 500 million people across 20 countries, spanning diverse cultural, educational, and technological contexts. In Spain and Latin America, LLM adoption is accelerating in universities, workplaces, and everyday life [30], yet research on LLM dependence remains scarce due to the absence of culturally and linguistically appropriate measurement tools. At present, the absence of a validated Spanish-language instrument limits the ability to assess the prevalence, structure and correlates of LLM dependency in Spanish-speaking populations and constrains participation in cross-cultural research that is increasingly important for LLM ethics and policy. Validating the LLM-D12 in Spanish is therefore an important methodological and conceptual step that may provide a foundation for developing evidence-informed guidelines for the integration of GenAI tools in education, mental health and professional training across the Spanish-speaking world and enable cross-cultural comparisons.

The present study aimed to examine the psychometric properties of the Spanish version of the LLM-D12 in a Spanish-speaking population. Specifically, the study assessed the internal structure, reliability, convergent and discriminant validity, and item-level functioning of the translated instrument. Additional analyses examined the network structure of items within each subscale and associations between LLM-D12 scores and patterns of LLM use as part of the validation process. In doing so, the study establishes a psychometrically validated Spanish-language instrument for assessing LLM dependency, supporting cross-cultural research in this area.

# 2. Methods

## *2.1. Data collection and sampling procedure*

*2.1.1. Data collection*. Data were collected using an online survey designed and administered via SurveyMonkey (www.surveymonkey.com). Participants were recruited through Prolific (www.prolific.com), a worldwide online platform for academic and commercial research that implements established data quality procedures. All participants remained anonymous, and digital informed consent was obtained prior to the start of the survey. Participants received monetary compensation upon successful completion of the survey, conditional on passing embedded attention checks.

To be eligible, participants had to be at least 18 years old, have Spanish origin and current Spanish residence. Before accessing the main questionnaire, participants indicated their level of engagement with the large language model (LLM) they used most frequently, which was designated as their “Primary LLM. Inclusion in the final sample required meeting at least one of the following three criteria: (a) almost daily use of the Primary LLM accompanied by perceived significant dependence; (b) almost daily use of the Primary LLM accompanied by perceived low dependence; or (c) perceived significant dependence on the Primary LLM even in the absence of daily use. Participants who reported neither regular use nor significant reliance on their Primary LLM were excluded from subsequent analyses. To ensure data quality, the survey included attention-check items that were carefully constructed to resemble the substantive scale items in tone, structure, and length, allowing inattentive responses to be identified without drawing undue attention to the checks.

*2.1.2. Sample size estimation*. Sample size planning for the Spanish-sample validation was informed by the model structure and parameter estimates obtained from the UK confirmatory factor analysis (12 items loading on two correlated factors, factor loadings ranging from .57 to .94). For the Spanish confirmatory factor analysis, the model comprised 12 observed variables loading on two correlated factors with no cross-loadings or correlated residuals, yielding 53 degrees of freedom. Using the RMSEA-based power approach [31], testing the null hypothesis of close fit (RMSEA = .05) against a poor-fitting alternative (RMSEA = .08) at α = .05, a sample size of 250 yields power of 0.88 for the test of close fit. Additional calculations indicated that a sample size of 220 is sufficient to estimate a reliability coefficient of α = .80 for a 12-item scale with a confidence interval width of ±.05. Accordingly, a target sample size of 250 was targeted to ensure adequately powered model-fit testing and precise estimation of reliability and correlational parameters in the Spanish validation.

*2.1.3. Sample in the present study*. Of the 473 responses collected, 3.38% failed attention check, 15.01% did not satisfy criteria of engagement with LLMs. These data were removed from analyses, yielding the final sample of 386 participants. The final sample had a mean age of 27.99 years (SD = 6.08); 55.18% of participants identified as male and 44.81% as female.

Most participants identified ChatGPT (OpenAI) as their Primary LLM (80.8%), followed by Gemini (7.3%), Copilot (4.1%), DeepSeek (3.6%). Other LLMs were occasionally reported but at substantially lower rates. Frequency data indicated that almost half of the sample (48.2%) interacted with LLMs multiple times per day.

*2.1.4. Spanish translation of the survey.* The Spanish version of the scale was developed using a standard back-translation procedure [32] to ensure conceptual and linguistic equivalence with the original instrument. First, the English version was translated into Spanish by a Spanish-English bilingual member of the research team fluent in both languages. This translation was then reviewed by a second bilingual researcher with expertise in human-AI interaction and research methods to improve clarity and accuracy. An independent bilingual academic, who was not involved in the study, had no prior familiarity with the scale, and had experience in survey design, subsequently translated the Spanish version back into English. Both the Spanish and the back-translated version was compared with the original scale by the senior author, and minor revisions were made to resolve discrepancies and clarify meaning.

### *2.2. Measures*

*2.2.1. Demographics*. Participants were asked to provide basic demographic information, including their age, gender, and educational background.

*2.2.2. Use of LLMs*. Participants were asked to identify their primary large language model and indicate how often they used it, using an 11-point scale ranging from 0 (“Very infrequently”) to 10 (“Very frequently”). They also reported the typical context of their LLM use, choosing whether they primarily interacted with it in stationary settings (e.g., home, office, school), on the move (e.g., commuting or traveling), or in both contexts. In addition, participants specified the device they most commonly used to access their LLM, with response options including PC/Computer, Mobile/Smartphone, or Other (e.g., tablet, smart speaker, dedicated AI Device). Participants also indicated their main mode of interaction with the LLM (text, voice, a combination of text and voice).

*2.2.3. LLM-D12-SP*. The LLM-D12-SP scale includes two subscales. Instrumental dependency, comprising 6 items, reflects a cognitive and habitual reliance on LLMs, such as memory, decision-making, and task automation. Relationship dependency, comprising 6 items, indicates users' parasocial and affective dependence users may develop towards LLMs. In line with the original LLM-D12 design, all items in the Spanish adaptation were assessed using a 6-point Likert scale ranging from 1 ("Strongly disagree") to 6 ("Strongly agree"), with no neutral midpoint. The absence of a midpoint was intended to encourage respondents to express a definite stance toward each statement. The full set of LLM-D12 items in both English and Spanish are presented in Appendix 1.

*2.2.4. Internet Addiction (IA).* Internet addiction was measured using the 7-item short version of the Internet Addiction Test (IAT-7) [33]. Example of items: How often do you neglect household chores to spend more time on-line? How often do you fear that life without the Internet would be boring, empty, and joyless? Responses were recorded on a 5-point Likert scale from 1 ("Never") to 5 ("Always"), yielding total scores ranging from 7 to 35. The Cronbach's α for this scale in our sample was 0.79, indicating a good internal reliability.

*2.2.5. Attitudes Toward AI (ATAI)*. Attitudes toward Artificial Intelligence [34] were assessed using the ATAI scale, which comprises two subscales: ATAI Acceptance (2 items) and ATAI Fear (3 items). Items were rated on an 11-point Likert scale ranging from 0 ("Strongly disagree") to 10 ("Strongly agree"). Possible scores ranged from 0 to 20 for the Acceptance subscale and from 0 to 30 for the Fear subscale. In the current sample, Cronbach's alpha was 0.67 for ATAI (Fear) and 0.74 for ATAI Acceptance.

*2.2.6. Need for Cognition (NFC)*. Need for Cognition was evaluated using the 6-item short version of Need for Cognition scale [35]. Example of items: I would prefer complex to simple problems; I like to have the responsibility of handling a situation that requires a lot of thinking. Participants responded on a 5-point Likert scale from 1 ("Extremely uncharacteristic of me") to 5 ("Extremely characteristic of me"), resulting in total scores between 6 and 30. In the present sample, the scale demonstrated high internal reliability, with a Cronbach's α of 0.87.

*2.2.7. Perceived Trustworthiness of primary LLM.* Trust in participants' primary LLM was measured using an 8-item scale [36]. The scale assesses eight core dimensions of trustworthiness: truthfulness, safety, fairness, robustness, privacy, machine ethics, transparency, and accountability. Example of items: I trust my primary LLM will provide information that is factually accurate and reliable, I trust my primary LLM will handle my personal data securely and confidentially. Items were rated on an 11-point Likert scale ranging from 0 ("Not at all") to 10 ("Completely"). Item scores were summed to yield a trust score ranging from 0 to 80. In the current sample, the scale exhibited strong internal reliability (Cronbach's α = 0.88).

### *2.3. Data analysis*

*2.3.1. Monotonicity check*. We applied nonparametric Mokken scale analysis, using the mokken package (version 3.1.2 in R) to assess monotonicity of participant responses. This method evaluates whether item endorsement probabilities increase consistently with higher levels of the underlying latent construct, a fundamental assumption for the validity of the measurement scales. All measures were without missing data across any of the items, and no item-level violations were identified. The scalability analyses indicated strong performance overall: the item scalability coefficients for the Instrumental Dependency items fell between 0.46 and 0.61, while those for the Relationship Dependency items ranged from 0.42 to 0.64. Every item surpassed the commonly accepted minimum criterion of 0.30, indicating adequate scalability throughout (Table 1).

**Table 1** Item scalability coefficients (Hi) for the Instrumental and Relational Dependency subscales of the LLM-D12-SP

| Instrumental Dependency | | Relationship Dependency | |
|---|---|---|---|
| *Item English/Spanish* | *Scalability Coefficient ($Hi)* | *Item English/Spanish* | *Scalability Coefficient ($Hi)* |
| **Without it, I feel less confident when making decisions** /Sin él, me siento menos seguro a la hora de tomar decisions | 0.61 | **I share details about my private life with it** /Comparto detalles sobre mi vida privada con él | 0.54 |
| **I use it sometimes without realizing how much time I spend immersed in it** /A veces lo uso sin darme cuenta de cuánto tiempo paso en él | 0.46 | **I interact with it as if it were a genuine companion** /Interactúo con él como si fuera un compañero de Verdad | 0.56 |
| **I feel much more rewarded and pleased when completing tasks using it** /Me siento mucho más satisfecho y contento cuando completo tareas utilizándolo | 0.5 | **It helps me feel less alone when I need to talk to someone /**Me ayuda a sentirme menos solo cuando necesito hablar con alguien | 0.64 |
| **I turn to it for support in decisions, even when I can make them myself with some effort**/Recurro a él en busca de apoyo para tomar decisiones, incluso cuando puedo tomarlas por mí mismo con un poco de esfuerzo | 0.55 | **It adds to my social life, making socializing more engaging and interesting** /Enriquece mi vida social, haciendo que las relaciones sociales sean más atractivas e interesantes | 0.57 |
| **It is my go-to for assistance in decision-making** /Es mi recurso habitual para obtener ayuda en la toma de decisions | 0.6 | **I use it solely as a tool, not to express my feelings or expect it to understand me [R]** /No espero que me entienda, ni comparto ni discuto mis sentimientos con él. Lo uso únicamente como una herramienta [R] | 0.42 |
| **Making decisions without it feels somewhat uneasy**/Tomar decisiones sin él resulta algo incómodo. | 0.55 | **It helps me feel less alone, reducing the need to talk to others** /Me ayuda a sentirme menos solo, reduciendo la necesidad de hablar con otras personas | 0.59 |

*Note.* Hi values indicate item scalability, with higher values reflecting stronger conformity to the latent trait. Interpretations follow conventional Mokken thresholds (Hi ≥ .40 = strong scalability). [R] indicates reverse-scored item

*2.3.2. Confirmatory Factor Analyis (CFA).* CFA was conducted to examine the proposed two-factor model, consisting of Instrumental Dependency and Relationship Dependency. Overall model adequacy was assessed using a range of fit statistics, including the Comparative Fit Index (CFI), Tucker-Lewis Index (TLI), Root Mean Square Error of Approximation (RMSEA) with its associated confidence interval, the Standardized Root Mean Square Residual (SRMR), and the chi-square test. Mahalanobis distance was employed to identify potential multivariate outliers. To evaluate how well each observed item represented its intended latent factor, standardized factor loadings and item-specific explained variance ($R^2$) values were computed.

*2.3.3. Discriminant validity.* Discriminant validity between the Instrumental Dependency and Relationship Dependency was evaluated using several complementary methods. First, the Fornell-Larcker criterion was employed by comparing each construct's average variance extracted (AVE) with the squared correlation between the two constructs: Discriminant validity is achieved when the AVE for each factor exceeds the shared variance. Second, the Heterotrait–Monotrait ratio of correlations (HTMT) was computed, which examines the extent to which correlations between items from different constructs are smaller than correlations among items within the same construct. HTMT values below the recommended cutoff of 0.85 were interpreted as evidence of adequate discriminant validity. Lastly, a chi-square difference test was used to contrast the hypothesized two-factor solution with a competing one-factor model in which all items loaded onto a single latent variable. Model comparisons were based on differences in $\chi^2$ statistics as well as information criteria (AIC and BIC) and RMSEA values, with lower AIC and BIC values and reduced RMSEA indicating superior model fit.

*2.3.4. Network Analysis.* A network analysis was employed using all 12 items from the Instrumental Dependency and Relationship Dependency subscales. The network was estimated using the EBICglasso algorithm, which applies regularization to partial correlations to yield a parsimonious and interpretable network structure. Item centrality indices were subsequently computed to quantify the relative importance and influence of each item within the network. All analyses were performed in R (version 4.3.1), with network estimation conducted via the *bootnet* package (estimateNetwork function). Correlations appropriate for ordinal data were calculated using *qgraph's cor_auto* function, and both network visualization and centrality metrics were generated using *qgraph*.

*2.3.5. External validation.* To test the external validity of the LLM-D12-SP, instrumental dependency and relationship dependency were analyzed simultaneously as outcome variables in a multivariate multiple regression model. Attitudes towards artificial intelligence (acceptance and fear), perceived trustworthiness of the primary large language model, internet addiction, and need for cognition were entered as predictors. The multivariate multiple regression model was estimated using ordinary least squares. Statistical significance of the predictor set was evaluated using multivariate test statistics derived from MANOVA, with Wilks' Lambda used as the primary test. Following the multivariate tests, two separate multiple linear regression models were estimated, one with instrumental dependency as the dependent variable and one with relationship dependency as the dependent variable, each including the same predictor set. For each regression, univariate ANOVA F tests, $R^2$, and adjusted $R^2$ were computed, and Cohen's $f^2$ was calculated as a measure of effect size. Ninety-five per cent confidence intervals for regression coefficients were obtained using the Wald method. Multicollinearity was assessed using generalized variance inflation factors (VIFs) for each univariate regression model. Finally, potential common-method variance was examined using Harman's single-factor test, implemented as a one-factor confirmatory factor analysis in which all LLM-D12 items were specified to load onto a single latent factor. Model fit was evaluated using the $\chi^2$ test, CFI, TLI, RMSEA, and SRMR. We also performed a sensitivity analysis to test whether the sample size in the present study was adequate to detect effects of the magnitude observed in both regression models for external validation.

*2.3.6. Additional analysis*. We also tested whether dependency on LLMs was associated with patterns of LLM use. Specifically, we examined whether frequency of LLM use (multiple times daily, once daily, 3-6 times per week, 1-2 times per week, 1-3 times per month), location of use (stationary, on the move, or both), device used (PC/computer, Mobile/Smartphone, Other) and interaction modality (Text, Voice, Text and Voice) were associated with instrumental dependency and relationship dependency, using total scores of the two subscales of the LLM-D12-SPI. We employed using a multivariate analysis of variance (MANOVA), with instrumental dependency and relationship dependency scores entered jointly as outcome variables and LLM usage predictors entered simultaneously. Multivariate effects were evaluated using Wilks' lambda. Following the multivariate test, separate multiple linear regression analyses were conducted for instrumental and relationship dependency to examine outcome-specific effects.

# 3. Results

## 3.1. Descriptive Analysis

Descriptive statistics the Instrumental Dependency and Relationship Dependency items are reported in Table 2.

Table 2
Mean and standard deviation for the Instrumental and Relationship Dependency Scale items

| **Instrumental Dependency** | | | **Relationship Dependency** | | |
|---|---|---|---|---|---|
| ***Item English/Spanish*** | ***Mean*** | ***SD*** | ***Item English/Spanish*** | ***Mean*** | ***SD*** |
| **Without it, I feel less confident when making decisions** /Sin él, me siento menos seguro a la hora de tomar decisions | 2.523 | 1.353 | **I share details about my private life with it** /Comparto detalles sobre mi vida privada con él | 2.896 | 1.613 |
| **I use it sometimes without realizing how much time I spend immersed in it** /A veces lo uso sin darme cuenta de cuánto tiempo paso en él | 2.552 | 1.492 | **I interact with it as if it were a genuine companion** /Interactúo con él como si fuera un compañero de Verdad | 2.562 | 1.583 |
| **I feel much more rewarded and pleased when completing tasks using it** /Me siento mucho más satisfecho y contento cuando completo tareas utilizándolo | 3.083 | 1.414 | **It helps me feel less alone when I need to talk to someone /**Me ayuda a sentirme menos solo cuando necesito hablar con alguien | 1.891 | 1.328 |
| **I turn to it for support in decisions, even when I can make them myself with some effort**/Recurro a él en busca de apoyo para tomar decisiones, incluso cuando puedo tomarlas por mí mismo con un poco de esfuerzo | 3.000 | 1.503 | **It adds to my social life, making socializing more engaging and interesting** /Enriquece mi vida social, haciendo que las relaciones sociales sean más atractivas e interesantes | 1.635 | 1.046 |
| **It is my go-to for assistance in decision-making** /Es mi recurso habitual para obtener ayuda en la toma de decisions | 2.881 | 1.467 | **I use it solely as a tool, not to express my feelings or expect it to understand me [R]** /No espero que me entienda, ni comparto ni discuto mis sentimientos con él. Lo uso únicamente como una herramienta [R] | 2.350 | 1.572 |
| **Making decisions without it feels somewhat uneasy**/Tomar decisiones sin él resulta algo incómodo. | 2.054 | 1.231 | **It helps me feel less alone, reducing the need to talk to others** /Me ayuda a sentirme menos solo, reduciendo la necesidad de hablar con otras personas | 1.684 | 1.083 |

To evaluate the accuracy of the items' means, 95% confidence intervals were computed for each item (Fig. 1).

## 3.2. Confirmatory factor analysis

The model demonstrated an overall acceptable fit to the data. The comparative fit index (CFI = 0.91) exceeded the conventional threshold of 0.90, and the standardized root mean square residual (SRMR = 0.07) was within recommended limits. The Tucker-Lewis index was slightly below the recommended cut-off (TLI = 0.89), and the root mean square error of approximation was elevated (RMSEA = 0.10, 90% CI [0.089, 0.114]).

However, RMSEA is known to be sensitive to model complexity and degrees of freedom, often yielding inflated values in models with relatively few degrees of freedom and a moderate number of indicators, even when other fit indices indicate acceptable model fit [37]. Simulation studies have shown that RMSEA can over-reject correctly specified models under such conditions, particularly in psychometric models with simple factor structures and correlated latent variables [38]. In this context, greater weight was placed on complementary fit indices less affected by model degrees of freedom, including the CFI and SRMR (Lai & Green, 2016), both of which fell within acceptable ranges.

Multivariate outliers were identified using Mahalanobis distance. Model fit was then evaluated both including and excluding these cases. Eight observations (2% of the sample) were flagged as potential outliers. Their exclusion produced negligible changes in model fit indices, factor loadings, and substantive conclusions; therefore, all results are reported for the full sample.

All items loaded significantly on their intended latent factors (all $p < .001$; see Table 3). Standardized factor loadings were generally strong and exceeded the conventional threshold of 0.50, with the exception of *Solely_as_a_tool* [R], which showed a slightly lower loading ($\lambda = 0.48$). For the Instrumental Dependency subscale, loadings ranged from 0.60 to 0.83, whereas loadings for the Relationship Dependency subscale ranged from 0.48 to 0.92. For Relationship Dependency items, explained variance was generally higher, with $R^2$ values ranging from 0.23 to 0.85. These results indicate that all items functioned as acceptable indicators of their respective latent constructs. Items with comparatively lower $R^2$ values (e.g., *More_rewarded* and *Solely_as_a_tool* [R]) appear to reflect additional item-specific variance not fully captured by the latent factors; however, their performance remained within acceptable bounds, supporting their retention in the measurement model.

Table 3
Standardised factor loadings and item-level explained variance for the LLM-D12-SP

| Scale | Item English/Spanish | Loadings | SE | z (p<.001) | Uniqueness | R2 |
|---|---|---|---|---|---|---|
| **Instrumental Dependency** | **Without it, I feel less confident when making decisions** /Sin él, me siento menos seguro a la hora de tomar decisions | 0.786 | 0.024 | 33.089 | 0.382 | 0.618 |
| | **I use it sometimes without realizing how much time I spend immersed in it** /A veces lo uso sin darme cuenta de cuánto tiempo paso en él | 0.597 | 0.036 | 16.483 | 0.643 | 0.357 |
| | **I feel much more rewarded and pleased when completing tasks using it** /Me siento mucho más satisfecho y contento cuando completo tareas utilizándolo | 0.628 | 0.034 | 18.292 | 0.605 | 0.395 |
| | **I turn to it for support in decisions, even when I can make them myself with some effort**/Recurro a él en busca de apoyo para tomar decisiones, incluso cuando puedo tomarlas por mí mismo con un poco de esfuerzo | 0.759 | 0.026 | 29.578 | 0.424 | 0.576 |
| | **It is my go-to for assistance in decision-making** /Es mi recurso habitual para obtener ayuda en la toma de decisions | 0.828 | 0.021 | 39.715 | 0.314 | 0.686 |
| | **Making decisions without it feels somewhat uneasy**/Tomar decisiones sin él resulta algo incómodo. | 0.716 | 0.029 | 24.999 | 0.487 | 0.513 |
| **Relationship Dependency** | **I share details about my private life with it** /Comparto detalles sobre mi vida privada con él | 0.573 | 0.036 | 15.717 | 0.671 | 0.329 |
| | **I interact with it as if it were a genuine companion** /Interactúo con él como si | 0.665 | 0.031 | 21.599 | 0.558 | 0.442 |

*Note*: SE = standard error. z values test whether factor loadings differ from zero; all loadings were statistically significant at $p < .001$. Uniqueness represents residual variance not explained by the latent factor ($1 - R^2$). $R^2$ indicates the proportion of variance in each item explained by its corresponding latent construct. [R] denotes a reverse-scored item

| Scale | Item English/Spanish | Loadings | SE | z (p<.001) | Uniqueness | R2 |
|---|---|---|---|---|---|---|
| | fuera un compañero de Verdad | | | | | |
| | **It helps me feel less alone when I need to talk to someone /**Me ayuda a sentirme menos solo cuando necesito hablar con alguien | 0.921 | 0.013 | 71.146 | 0.152 | 0.848 |
| | **It adds to my social life, making socializing more engaging and interesting** /Enriquece mi vida social, haciendo que las relaciones sociales sean más atractivas e interesantes | 0.790 | 0.022 | 35.955 | 0.376 | 0.624 |
| | **I use it solely as a tool, not to express my feelings or expect it to understand me [R]** /No espero que me entienda, ni comparto ni discuto mis sentimientos con él. Lo uso únicamente como una herramienta [R] | 0.482 | 0.041 | 11.664 | 0.768 | 0.232 |
| | **It helps me feel less alone, reducing the need to talk to others** /Me ayuda a sentirme menos solo, reduciendo la necesidad de hablar con otras personas | 0.841 | 0.018 | 46.232 | 0.293 | 0.707 |

*Note*: SE = standard error. z values test whether factor loadings differ from zero; all loadings were statistically significant at $p < .001$. Uniqueness represents residual variance not explained by the latent factor ($1 - R^2$). $R^2$ indicates the proportion of variance in each item explained by its corresponding latent construct. [R] denotes a reverse-scored item

## 3.3. Validation

*3.3.1. Internal consistency.* An overall Cronbach's α of LLM-D12-SP was 0.89, indicating good internal coherence. Both subscales showed strong internal consistency (α = 0.86 and α = 0.85). suggesting that both scales and its combined scale possess sufficient reliability for both empirical research and practical applications. For the two subscales, all corrected item-total correlations (r.cor) exceeded the commonly accepted minimum criterion of 0.30 (Table 4), indicating that each item was meaningfully and positively associated with the composite scale score. Corrected item-total correlations for the Instrumental Dependency scale fell between 0.57 and 0.73. For the Relationship Dependency scale, corrected item-total correlations ranged from 0.45 to 0.76, indicating that all items contribute substantially to the assessment of the latent construct and show strong alignment with the overall scale score.

Table 4
Internal consistency and item analysis indices for Instrumental Dependency and Relationship Dependency

| Subscale | Item English/Spanish | Chronbach's α | α SE | r.cor | var.r | r.drop |
|---|---|---|---|---|---|---|
| **Instrumental Dependency** | **Without it, I feel less confident when making decisions** /Sin él, me siento menos seguro a la hora de tomar decisions | 0.879 | 0.0092 | 0.68 | 0.017 | 0.63 |
| | **I use it sometimes without realizing how much time I spend immersed in it** /A veces lo uso sin darme cuenta de cuánto tiempo paso en él | 0.883 | 0.0089 | 0.6 | 0.019 | 0.56 |
| | **I feel much more rewarded and pleased when completing tasks using it** /Me siento mucho más satisfecho y contento cuando completo tareas utilizándolo | 0.884 | 0.0088 | 0.57 | 0.018 | 0.53 |
| | **I turn to it for support in decisions, even when I can make them myself with some effort**/Recurro a él en busca de apoyo para tomar decisiones, incluso cuando puedo tomarlas por mí mismo con un poco de esfuerzo | 0.878 | 0.0093 | 0.68 | 0.018 | 0.64 |
| | **It is my go-to for assistance in decision-making** /Es mi recurso habitual para obtener ayuda en la toma de decisions | 0.875 | 0.0095 | 0.73 | 0.017 | 0.69 |
| | **Making decisions without it feels somewhat uneasy**/Tomar decisiones sin él resulta algo incómodo. | 0.881 | 0.0091 | 0.63 | 0.018 | 0.59 |
| **Relationship Dependency** | **I share details about my private life with it** /Comparto detalles sobre mi vida privada con él | 0.882 | 0.0090 | 0.62 | 0.019 | 0.59 |
| | **I interact with it as if it were a genuine companion** /Interactúo con él como si fuera un compañero de Verdad | 0.879 | 0.0092 | 0.66 | 0.019 | 0.62 |
| | **It helps me feel less alone when I need to talk to someone /**Me ayuda a sentirme menos solo cuando necesito hablar con alguien | 0.876 | 0.0093 | 0.76 | 0.015 | 0.68 |

*Note*: α SE represents the standard error of the Cronbach's α estimate. R.cor – the item-total correlation after correcting for item overlap (i.e., the correlation between the item and the total of the remaining items, not including itself). var.r – the variance of the item-item correlations with low values suggests item homogeneity. r.drop – corrected item-total correlation if the item dropped

| Subscale | Item English/Spanish | Chronbach's α | α SE | r.cor | var.r | r.drop |
|---|---|---|---|---|---|---|
| | **It adds to my social life, making socializing more engaging and interesting** /Enriquece mi vida social, haciendo que las relaciones sociales sean más atractivas e interesantes | 0.879 | 0.0092 | 0.73 | 0.017 | 0.67 |
| | **I use it solely as a tool, not to express my feelings or expect it to understand me [R]** /No espero que me entienda, ni comparto ni discuto mis sentimientos con él. Lo uso únicamente como una herramienta [R] | 0.891 | 0.0083 | 0.45 | 0.016 | 0.42 |
| | **It helps me feel less alone, reducing the need to talk to others** /Me ayuda a sentirme menos solo, reduciendo la necesidad de hablar con otras personas | 0.880 | 0.0091 | 0.7 | 0.016 | 0.63 |

*Note*: α SE represents the standard error of the Cronbach's α estimate. R.cor – the item-total correlation after correcting for item overlap (i.e., the correlation between the item and the total of the remaining items, not including itself). var.r – the variance of the item-item correlations with low values suggests item homogeneity. r.drop – corrected item-total correlation if the item dropped

*3.3.2. Discriminant validity.* Discriminant validity between the Instrumental and Relationship Dependency subscales was assessed using multiple complementary approaches. According to the Fornell-Larcker criterion, the average variance extracted (AVE) was 0.52 for Instrumental Dependency and 0.48 for Relationship Dependency, both exceeding the squared correlation between the two latent factors (0.32). This pattern supports adequate discriminant validity. Consistent with this result, the heterotrait–monotrait ratio (HTMT) was 0.65, well below the recommended threshold of 0.85, indicating that the two constructs were empirically distinct. Further evidence was provided by model comparison analyses. A chi-square difference test showed that the two-factor model fit the data significantly better than a one-factor alternative in which all items loaded onto a single latent construct ($\Delta\chi^2 = 488$, $p < .001$). In addition, the two-factor model demonstrated superior information criteria (AIC = 13,974.40; BIC = 14,073.30) compared with the one-factor model (AIC = 14,460.39; BIC = 14,555.33), further supporting the distinction between Instrumental and Relationship Dependency.

*3.3.3. Network analysis.* Within the Instrumental Dependency network, degree centrality indicated that *It_is_my_go_to* (degree = 1.13) and *Without_it_less_confident* (degree = 1.10) were the most strongly connected items. In terms of closeness centrality, which reflects how efficiently an item is connected to all other nodes in the network, the highest values were observed for *Immersed_in_it* (closeness = 0.0094), *Decisions_without_uneasy* (closeness = 0.0093), and *Turn_to_it_even_when_able* (closeness = 0.0093), placing these items among the more central nodes in the Instrumental Dependency network. For the Relationship Dependency network, *Feel_less_alone* (degree = 1.42) and *It_adds_to_my_social_life* (degree = 1.06) showed the highest degree centrality. These items also exhibited relatively high closeness centrality (*Feel_less_alone*: 0.010; *It_adds_to_my_social_life*: 0.015), indicating that they were well connected within the network structure. Together, these results suggest that items reflecting perceived social support and enhanced social engagement were structurally central within the Relationship Dependency subscale. Overall, the network analysis indicated that items clustered into two internally coherent and structurally distinct networks corresponding to Instrumental and Relationship Dependency (Fig. 1).

*3.3.4. External Validation.* A multivariate multiple regression analysis was conducted to examine whether AI Acceptance, AI Fear, Internet Addiction, Need for Cognition, and perceived trustworthiness of LLMs were associated with scores on the Instrumental Dependency and Relationship Dependency subscales. The overall multivariate test was significant (Wilks' $\Lambda = 0.704$, $F(10, 758) = 14.529$, $p < .001$), indicating that the predictors, considered jointly, were related to the combined set of dependency scores.

For Instrumental Dependency, the model explained 25.5% of the variance ($R^2 = 0.264$, adjusted $R^2 = 0.255$), $F(5, 380) = 27.32$, $p < .001$. Higher scores on AI Acceptance ($\beta = 0.19$, $SE = 0.06$, $p < .05$), Internet Addiction ($\beta = 0.25$, $SE = 0.05$, $p < .001$), and perceived trustworthiness of LLMs ($\beta = 0.28$, $SE = 0.06$, $p < .001$) were associated with higher Instrumental Dependency scores. For AI Fear, the association did not meet the conventional $p < .05$ criterion ($\beta = 0.10$, $SE = 0.05$, $p < .10$). For Relationship Dependency, the model accounted for 17.1% of the variance ($R^2 = 0.182$, adjusted $R^2 = 0.171$), $F(5, 380) = 16.90$, $p < .001$. Higher Relationship Dependency scores were associated with greater AI Acceptance ($\beta = 0.13$, $SE = 0.04$, $p < .05$), Internet Addiction ($\beta = 0.265$, $SE = 0.05$, $p < .001$), and perceived trustworthiness of LLMs ($\beta = 0.21$, $SE = 0.06$, $p < .001$) (see Table 5).

Table 5
Univariate linear regression models for Instrumental and Relationship Dependency scores

| Outcome | Predictor | Estimate | SE | t-test | p.value |
|---|---|---|---|---|---|
| Instrumental Dependency | **AI Acceptance** | **0.193** | **0.059** | **3.298** | **0.001** |
| | **AI Fear** | **0.102** | **0.049** | **2.096** | **0.037** |
| | NFC | -0.039 | 0.045 | -0.861 | 0.390 |
| | **Internet Addiction** | **0.253** | **0.046** | **5.556** | **0.000** |
| | **Perceived Trustworthiness of LLM** | **0.279** | **0.056** | **5.032** | **0.000** |
| Relationship Dependency | **AI Acceptance** | **0.128** | **0.062** | **2.078** | **0.038** |
| | AI Fear | 0.058 | 0.051 | 1.132 | 0.258 |
| | NFC | -0.028 | 0.048 | -0.576 | 0.565 |
| | **Internet Addiction** | **0.265** | **0.048** | **5.507** | **0.000** |
| | **Perceived Trustworthiness of LLM** | **0.214** | **0.059** | **3.654** | **0.000** |

*Note*: Estimate represents the standardized regression coefficient (β), indicating the direction and strength of the association between the predictor and the outcome, controlling for all other predictors. SE - the standard error of the estimate

Post-hoc Wald *z* tests were conducted to compare the regression coefficients for each predictor across the two outcome variables, Instrumental Dependency and Relationship Dependency. These analyses indicated that none of the predictors differed significantly in their effects between the two subscales, suggesting that the associations of each predictor were comparable across both dependency measures (see Table 6).

Table 6
Wald test comparing predictor effects across Instrumental and Relationship Dependency

| Predictor | Instrumental | Relationship | Diff | SE Diff | z | p | CI Lower | CI Upper |
|---|---|---|---|---|---|---|---|---|
| AI Acceptance | 0.193 | 0.128 | 0.065 | 0.085 | 0.762 | 0.446 | -0.102 | 0.232 |
| AI Fear | 0.102 | 0.058 | 0.044 | 0.071 | 0.620 | 0.535 | -0.095 | 0.182 |
| NFC | 0.253 | 0.265 | -0.011 | 0.066 | -0.174 | 0.862 | -0.141 | 0.118 |
| Internet Addiction | -0.039 | -0.028 | -0.012 | 0.066 | -0.175 | 0.861 | -0.141 | 0.118 |
| Perceived Trustworthiness of LLM | 0.279 | 0.214 | 0.065 | 0.081 | 0.811 | 0.417 | -0.093 | 0.224 |

*Note*: Diff represents the difference between the Instrumental and Relationship coefficients; SE Diff is the standard error of the difference; z is the Wald test statistic for the difference; p is the p-value testing whether the difference is significantly different from zero; CI l and CI up represent the lower and upper bounds of the 95% confidence interval for the difference.

Following the external validation analyses, a sensitivity analysis was conducted to assess whether the available sample size was adequate given the observed effect sizes in the multivariate multiple regression models. For the Instrumental Dependency model (adjusted $R^2 = 0.255$; five predictors; $N = 386$), the corresponding Cohen's $f^2$ was 0.36, which falls within the range conventionally described as a large effect. For the Relationship Dependency model (adjusted $R^2 = 0.171$; five predictors; $N = 386$), the resulting $f^2$ was 0.22, indicating a medium-to-large effect. These results indicate that the sample size used in the external validation analyses was sufficient to detect effects of the magnitude observed in both models.

## 3.4. Additional Analysis

We examined whether dependency on LLMs was associated with LLM usage using two multiple regression models. In both models, frequency of LLM use, location of use, categorical frequency of use, device used, and interaction method were entered simultaneously as predictors. Categories with small cell counts were combined prior to analysis. Location of use category “on the move” was merged with “both stationary and on the move” into a single category labelled “both stationary and on the move”. For frequency of use, “1–2 times per week”, “1–3 times per month”, and “Less than once per month” were combined into “Less than daily”. For device used, the category “Other” (e.g., tablet, smart speaker, dedicated AI Device, etc.) was selected only by one participant and we removed it from the analysis. For interaction method, “Voice” and “Text and Voice” were combined into “Text and Voice” (Table 7).

Table 7
Descriptive statistics for the LLM usage

| Usage | Categories | N | Proportion | Instrumental Dependency | | Relationship Dependency | |
|---|---|---|---|---|---|---|---|
| | | | | Mean | SD | Mean | SD |
| **Device** | PC / Computer | 249 | 64.5 | 15.5 | 6 | 14.8 | 6.38 |
| | Mobile / Smartphone | 136 | 35.2 | 17.2 | 6.7 | 12.1 | 6.05 |
| **Frequency of use** | Multiple times daily | 186 | 48.2 | 18 | 6.7 | 14.5 | 6.5 |
| | Once daily | 72 | 18.7 | 15.7 | 6.5 | 12 | 6.1 |
| | Less than daily | 127 | 32.9 | 13.5 | 5.4 | 11.4 | 5.7 |
| **Interaction method** | Text | 336 | 87 | 15.6 | 6.5 | 12.7 | 6.3 |
| | Text and Voice | 50 | 13 | 19 | 6.2 | 15.5 | 5.9 |
| **Location of use** | Stationary | 224 | 58 | 14.8 | 6.2 | 11.9 | 6.2 |
| | Both stationary and on the move | 161 | 41.7 | 17.8 | 6.6 | 14.6 | 6.2 |

The multivariate analysis indicated significant effects of frequency of LLM use (Wilks' $\lambda = .839$, $F(2, 375) = 35.88$, $p < .001$), device used (Wilks' $\lambda = .943$, $F(2, 375) = 11.27$, $p < .001$), frequency of use (Wilks' $\lambda = .974$, $F(4, 750) = 2.46$, $p = .044$), and location of use (Wilks' $\lambda = .982$, $F(2, 375) = 3.48$, $p = .032$) on the combined dependency outcomes. Interaction method was not significant at the multivariate level (Wilks' $\lambda = .994$, $F(2, 375) = 1.12$, $p = .329$). These findings indicate that frequency of use, device, and location are associated with overall dependency when Instrumental and Relationship Dependency are considered jointly.

The regression model predicting Instrumental Dependency was significant, $F(6, 376) = 14.22$, $p < .001$, explaining 18.5% of the variance ($R^2 = .185$, adjusted $R^2 = .172$). Frequency of LLM use was positively associated with Instrumental Dependency ($b = 1.072$, $SE = 0.230$, $t = 4.663$, $p < .001$). Compared with less-than-daily use, multiple-times-daily use was associated with higher instrumental dependency ($b = 1.824$, $SE = 0.841$, $t = 2.169$, $p = .031$), whereas once-daily use did not differ ($b = 0.633$, $SE = 0.918$, $t = 0.689$, $p = .491$). Participants using LLMs only in stationary settings reported lower Instrumental Dependency than those using LLMs both stationary and on the move ($b = -1.570$, $SE = 0.670$, $t = -2.342$, $p = .020$). Device used ($b = -1.079$, $SE = 0.690$, $p = .119$) and interaction method ($b = 1.415$, $SE = 0.960$, $p = .141$) were not associated with instrumental dependency

The model predicting Relationship Dependency was significant, $F(6, 376) = 12.22$, $p < .001$, accounting for 16.3% of the variance ($R^2 = .163$, adjusted $R^2 = .150$). Higher frequency of LLM use was linked to greater Relationship Dependency ($b = 0.994$, $SE = 0.224$, $t = 4.427$, $p < .001$). In contrast, frequency of use categories did not differentiate Relationship Dependency: neither once-daily use ($b = -0.959$, $SE = 0.896$, $t = -1.070$, $p = .285$) nor multiple times daily use ($b = 0.692$, $SE = 0.821$, $t = 0.843$, $p = .400$) differed from less than daily use. Device type emerged as a significant predictor, with PC/computer users reporting lower Relationship

Dependency than mobile/smartphone users (b = -2.464, SE = 0.673, t = -3.659, p < .001). Location of use was not significantly associated with relationship dependency (stationary-only vs both stationary and on the move: b = -1.189, SE = 0.654, t = -1.818, p = .070). Interaction method was also not significantly associated with Relationship Dependency (text and voice vs text: b = 0.859, SE = 0.937, t = 0.916, p = .360).

## 4. Discussion

# 4.1. Psychometric properties of LLM-D12-SP scale

The present study examined the psychometric properties of the Spanish version of the LLM-D12 as a measure of users' Instrumental Dependency and Relationship Dependency on their primary large language model. The confirmatory factor analysis indicated that the original two-subscale structure was retained, with items grouping into Instrumental Dependency and Relationship Dependency as intended. The two-subscale solution provided a better representation of the data than a single-factor model, indicating that the two forms of dependency were distinguished empirically. This support analyzing the subscales separately when assessing LLM dependency in Spanish-speaking samples. Both subscales showed high internal consistency, with reliability estimates exceeding commonly accepted thresholds, indicating that the translated items operate consistently in the Spanish version.

At the item level, all indicators loaded on their intended subscales. The reverse-scored item *Solely_as_a_tool* showed the weakest loading within the Relationship Dependency subscale. This is consistent with established psychometric findings showing that reverse-worded items often perform less strongly due to increased processing demands or inconsistent interpretation [39][40]. Despite this, the item remains conceptually important because it reflects resistance to viewing the LLM purely as a functional tool. Given the performance of the remaining items and the overall reliability of the subscale, its inclusion does not undermine the validity of the measure. Evidence from external validation analyses further supports the construct validity of the LLM-D12-SP. Scores on both dependency dimensions were associated with external measures in directions consistent with existing research on technology-related dependency, indicating that the scale captures a form of reliance that aligns with established dependency constructs.

Higher LLM dependency was associated with broader indicators of problematic and habitual internet use. This finding is consistent with prior research showing that reliance on specific digital systems often appears alongside more general tendencies towards technology-related dependency [41][42]. From a measurement perspective, this supports convergent validity, as the LLM-D12-SP is linked to related constructs while remaining focused on dependency specific to LLM use.

Perceived trustworthiness of LLMs was positively associated with both Instrumental Dependency and Relationship Dependency. This is consistent with human-automation research showing that trust is a central driver of reliance on automated systems because it supports willingness to delegate, accept assistance, and continue use [43][44]. Applied to LLMs, users who judge the system as reliable and competent are more likely to use it as a dependable source of support, both for practical tasks and for interactional needs. The fact that trustworthiness relates to both dependency domains provides convergent validity evidence for the LLM-D12-SP and supports the interpretation that higher dependency scores reflect confidence-based reliance on LLMs.

Attitudes towards AI further clarify the construct captured by the LLM-D12-SP. Higher dependency scores were associated with more positive attitudes towards AI, whereas fear-based attitudes showed little association. This pattern is consistent with technology acceptance models [45], which emphasize perceived usefulness, acceptance, and perceived benefits as key correlates of sustained engagement with technological systems, while anxiety and perceived threat play a more limited role once use is established [46]. From a validation perspective, the divergence between positive and fear-related attitudes is informative: it indicates that higher dependency is aligned with favorable evaluations of AI, while remaining largely independent of apprehension or threat perceptions. This provides discriminant validity evidence that the scale captures reliance grounded in acceptance, not general anxiety about AI.

Finally, the lack of association between LLM dependency and Need for Cognition supports discriminant validity and further differentiates LLM dependency from broader patterns of problematic internet use. Need for Cognition reflects a dispositional preference for effortful, self-directed thinking [47] and has been shown to correlate negatively with Internet Addiction, where reliance on online environments may function as a substitute for sustained cognitive engagement [48]. In contrast, reliance on efficient external cognitive tools does not appear to be systematically associated with cognitive style, particularly when those tools are perceived as instrumental and timesaving [49]. The absence of a relationship observed here indicates that the LLM-D12-SP is not indexing general cognitive motivation or avoidance of effort, but a more specific tendency to depend on LLMs for instrumental and relationship purposes.

Taken together, these results indicate that the Spanish version of the LLM-D12 functions as a coherent and interpretable psychometric measure, retaining its intended structure and demonstrating appropriate reliability and construct validity.

## 4.2. Comparison of Spanish, UK, and Arab validation of LLM-D12

Compared with the UK [36] and Arab [29] validations, the Spanish version of the LLM-D12 shows similar psychometric properties. As in both previous studies, the Spanish data support a two-subscale structure distinguishing Instrumental Dependency and Relationship Dependency, with the two dimensions remaining empirically separable and clearly preferable to a single-factor solution. This convergence across samples indicates that the basic measurement structure of the LLM-D12 is preserved in the Spanish adaptation.

Item-level behavior in the Spanish version closely mirrors that observed in the UK and Arab samples. Items load on their intended subscales in all three validations, with Relationship Dependency items generally showing stronger loadings and higher explained variance than Instrumental Dependency items. The reverse-scored item (*Solely_as_a_tool*) displays weaker loadings in the Spanish sample, consistent with its performance in the UK and Arab versions, suggesting that this reflects a stable item-level characteristic rather than a translation-specific issue.

Reliability estimates for the Spanish subscales are comparable to those reported in the UK and Arab validations. In the Spanish sample, internal consistency was high for both Instrumental Dependency ($\alpha$ = .86) and Relationship Dependency ($\alpha$ = .85). Similar values were observed in the UK study, where Cronbach's $\alpha$ was .84 for Instrumental Dependency and $\alpha$ = .91 for Relationship Dependency, and in the Arab validation, where reliability estimates were $\alpha$ = .85 for Instrumental Dependency and $\alpha$ = .90 for Relationship Dependency. Across all three samples, Relationship Dependency consistently showed equal or higher internal consistency than Instrumental Dependency, indicating a stable difference in internal coherence between the two subscales across languages.

Evidence for discriminant validity in the Spanish validation is also consistent with that reported in the UK and Arab studies. In the Spanish sample, the AVE was 0.52 for Instrumental Dependency and 0.48 for Relationship Dependency, both exceeding the squared correlation between the two factors (0.32), and the HTMT value was 0.65. Similar results were observed in the UK validation, where AVE values were 0.50 for Instrumental Dependency and 0.66 for Relationship Dependency, with an HTMT of 0.59. In the Arab validation, AVE values likewise exceeded the squared inter-factor correlation, and the HTMT value was 0.64. Across all three studies, these variance-based indices indicate that Instrumental Dependency and Relationship Dependency are empirically distinct constructs, supporting their use as non-redundant dimensions across linguistic contexts

Finally, the Spanish external validation results show a similar overall structure to those reported in the UK and Arab samples. In each case, dependency scores are associated with Internet Addiction and perceived trustworthiness of LLMs, while Need for Cognition shows weak or no association. These shared associations support the view that the Spanish version captures the same dependency-related tendencies as the original and Arab adaptations.

## 4.3. Theoretical and practical implications

This LLM-D12-SP validation contributes to theory by supporting the conceptual organization of dependency on LLMs as a multidimensional construct. The replication of the two-factor structure in a Spanish-speaking sample indicates that Instrumental Dependency and Relationship Dependency represent stable dimensions of reliance, extending beyond the linguistic and cultural context in which the scale was originally developed. This supports the theoretical position that dependency on LLMs cannot be adequately captured as a single continuum of use or engagement but instead reflects qualitatively different modes of reliance embedded in interaction with the system [10]. At the same time, variation in the strength of some relational indicators across language versions indicates an important theoretical boundary. While the underlying relational dependency construct remains intact, items tapping emotional disclosure and perceived intimacy (i.e., sharing private life) show comparatively weaker associations in the Spanish sample. This suggests that the relational dimension of LLM dependency is more sensitive to contextual interpretation at the item level, even when its overall structure is preserved. From a theoretical standpoint, this indicates that relational dependency is not defined by specific expressions of emotional closeness, but by a broader orientation towards experiencing the LLM as socially responsive.

The Spanish validation of the LLM-D12 has several implications for the use and interpretation of the scale in research settings including studies examining AI use in organizational and work-related contexts. First, the results support scoring the instrument using two subscales, Instrumental Dependency and Relationship Dependency. Using these two subscales allows researchers to distinguish between different forms of reliance on LLMs and avoids conflating conceptually distinct aspects of dependency, thereby enabling more precise examination of psychological processes underlying human–AI interaction.

Second, the consistency of item functioning across language versions indicates that the core indicators of both subscales operate reliably in Spanish. Items capturing routine reliance on LLMs for decision support and confidence, as well as items reflecting perceived social support from the system, contribute strongly to their respective subscales. However, some variability in item performance should be taken into account when interpreting results. In line with findings from other language versions, the reverse-scored item Solely_as_a_tool shows weaker associations with the Relationship Dependency factor. This suggests that responses to this item may be influenced by factors unrelated to the core construct, such as response format or item framing. Researchers should therefore interpret this item cautiously at the item level, while noting that its inclusion does not compromise the reliability or validity of the subscale as a whole. Therefore, interpretation of the Relationship Dependency subscale should be based on the overall score rather than on individual items. Differences in endorsement of specific relational statements do not necessarily imply absence of relational dependency but may reflect variation in how users express socially oriented engagement with LLMs. Using the subscale score as intended provides a more stable and interpretable measure of this form of dependency. Appendix 2 provides mapping of individual items of LLM-D12 into underlying theoretical constructs and measurements.

Third, the usage findings in the Spanish sample indicate that stronger dependency is associated with more frequent engagement with LLMs. Higher overall frequency of use was related to higher levels of both Instrumental and Relationship Dependency. When frequency was examined in categories, 'multiple times daily' users reported higher Instrumental Dependency than 'less than daily' users, whereas Relationship Dependency did not differ across these categories. In addition, individuals who reported using LLMs both in stationary settings and while on the move showed higher Instrumental Dependency than those using them only in stationary settings. Relationship Dependency, in contrast, differed by device: users who primarily accessed LLMs via mobile devices reported higher relational dependence than those using a PC. Overall, these pattern suggests that more frequent and more integrated use of LLMs is associated with stronger dependency, and this general pattern is consistent with the UK [11] and Arab [29] validations. Such findings are particularly relevant in contexts where LLMs are embedded into routine task performance and decision-making, including professional and organizational environments.

For researchers studying human-AI interaction, the scale provides a way to distinguish between reliance on LLMs for decision-making and reliance linked to socially oriented engagement. This allows studies to move beyond general usage metrics and to examine how different forms of dependency relate to design features, interaction styles, or user characteristics, supporting evaluation of reliance patterns and informing responsible AI design and implementation. For psychologists, organizational psychologists, and behavioural researchers, the scale offers a tool for examining dependency on LLMs as a specific technology-related construct. Its validated associations with patterns of internet use and trust in AI make it suitable for research on technology reliance, digital behavior, and related outcomes in Spanish-speaking samples, including research on AI use in organizational contexts and applied evaluation of AI-supported work practices.

## 4.4. Limitations and future directions

This study should be interpreted in light of several limitations. First, although the sample size was adequate for the psychometric analyses conducted, participants were recruited through an online, self-selected process. This may have resulted in a sample that is more comfortable with digital technologies and more engaged with LLMs than the broader Spanish population. As a result, the findings may not fully generalize to individuals with lower levels of exposure to, or interest in, LLM-based systems.

Second, all data were collected using self-report measures. While appropriate for scale validation, self-report data are susceptible to response biases, situational influences, and individual differences in interpretation. These factors may affect how dependency is reported, independent of actual patterns of LLM use. Future research would benefit from supplementing self-report with behavioral indicators, such as logged usage data, task reliance measures, or indicators of how LLMs are integrated into academic, occupational, or everyday activities.

Third, the cross-sectional design limits conclusions about stability or change in dependency over time. Reliance on LLMs is likely to evolve alongside rapid technological development, changes in system capabilities, and shifting regulatory or social norms. Longitudinal research is needed to examine how instrumental and relational dependency develop, persist, or decline, and how these trajectories relate to changes in use contexts and system design.

Finally, dependency on large language models remains a developing construct. Although the LLM-D12 is grounded in existing conceptual work, it should be viewed as an initial operationalization rather than a definitive framework. Current literature often uses terms such as dependency, overuse, overreliance, and addiction interchangeably, reflecting ongoing conceptual ambiguity. At present, there is limited evidence linking LLM dependency to negative outcomes in wellbeing, cognition, or social functioning. Interpreting dependency as inherently problematic would therefore be premature. By operationalizing dependency into measurable components, this study provides a foundation for future research to examine under what conditions, and for whom, reliance on LLMs may become maladaptive.

# 5. Conclusion

The Large Language Model Dependency Scale - Spanish version (LLM-D12-SP) is the first validated instrument for assessing dependency on large language models in Spanish-speaking LLM users. The scale demonstrates good psychometric properties and provides a reliable measure of LLM dependency that is suitable for use in research contexts involving Spanish-speaking populations. The LLM-D12-SP consists of two empirically distinct subscales, Instrumental Dependency and Relationship Dependency. Instrumental Dependency captures dependence on LLMs for decision-making, confidence, and cognitive support, whereas Relationship Dependency reflects socially and emotionally oriented dependence, including perceived companionship and social substitution. The two subscales show strong internal consistency and good discriminant validity, supporting their use as separate but related dimensions of LLM dependency. As LLM use continues to expand across Spanish-speaking countries, the LLM-D12-SP offers researchers and practitioners a validated Spanish-language tool for examining how different forms of dependency on LLMs manifest in these contexts. The availability of this instrument enables cross-cultural comparison with existing language versions and supports systematic investigation of LLM dependency in Spanish-speaking population across educational, occupational, and LLM design domains.

# Declarations

**Author Contributions**:

TGB: Performed and reported the statistical analyses; contributed to the first draft.

ME: Mentored the research; validated the analyses; reviewed and edited the paper.

SA: Curated the data; validated the analyses; contributed to the first draft; reviewed and edited the paper.

AGC: Conducted the Spanish translation; collected part of the data; reviewed the paper.

RA: Mentored the research; curated the data; validated the analyses; reviewed and edited the paper.

AY: Mentored the research; designed and validated the analyses; contributed to the first draft; reviewed and edited the paper.

**Funding**

Open Access Fund has been provided by Qatar National Library. This publication was supported by NPRP-14-Cluster Grant # NPRP 14 C- 0916–210015 from the Qatar National Research Fund (a member of the Qatar Foundation). The findings herein reflect the work and are solely the authors responsibility.

**Data availability**

The study design, dataset, and analysis files are available on the Open Science Framework (OSF) at the following link: https://osf.io/3b7ea/overview?view_only=8236ec964aed4e9ba127886fe00a4912 The author confirms that all data generated or analyzed during this study are included in this published article.

**Competing interest**

The authors declare no competing financial and/or non-financial interests.

**Accordance Statement**

Ethical approval for this research was granted by the Bournemouth University Ethics Committee, UK in accordance with the guidelines and regulations set forth by the Declaration of Helsinki [version 2024]. The protocol number for this approval is N62239, 03 March 2025. Informed consent was obtained from all individual participants included in the study.

### Consent to participate

Informed consent was obtained from all individual participants included in the study.

### Consent to publish

Not Applicable

# Figures

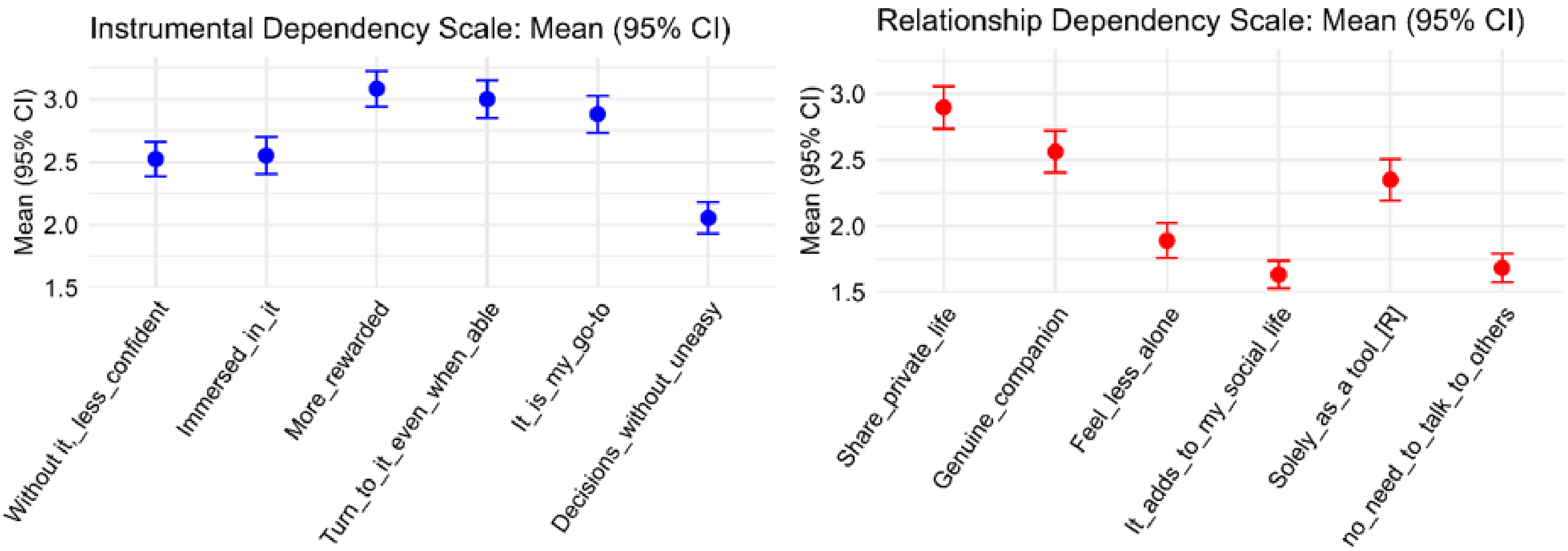

**Figure 1**

Item means with 95% confidence intervals for the Instrumental and Relationship Dependency scales

# Supplementary Files

This is a list of supplementary files associated with this preprint. Click to download.

- LargeLanguageModelDependencyQuestionnaire.docx